\documentclass[conference]{ieeeconf}
\IEEEoverridecommandlockouts 

\usepackage{cite}
\usepackage[pdftex]{graphicx}
    \graphicspath{{img/},{fig/}}
\usepackage{amsmath,amssymb}
\usepackage{array}
\usepackage[caption=false,font=footnotesize]{subfig}
\usepackage{fixltx2e}
\usepackage{siunitx}
\usepackage{threeparttable}

\usepackage{hyperref}
\usepackage[utf8]{inputenc}
\usepackage{gensymb}

\usepackage{xcolor}
\usepackage{soul}
\usepackage{float}

\makeatletter
\newcommand*\titleheader[1]{\gdef\@titleheader{#1}}
\AtBeginDocument{%
  \let\st@red@title\@title
  \def\@title{%
    \bgroup\normalfont\small\raggedright\@titleheader\par\egroup
    \vskip1.5em\st@red@title}
}
\makeatother

\titleheader{2022 IEEE International Conference on Robotics and Automation (ICRA 2022). Preprint version. Accepted February 2022.$^{2}$} 
\title{\LARGE \bf
Multimodal Hydrostatic Actuators for Wearable Robots: A Preliminary Assessment of Mass-Saving and Energy-Efficiency Opportunities
}

\author{Jeff Denis$^{1}$, Alex Lecavalier$^{1}$, Jean-Sébastien Plante$^{1}$ and Alexandre Girard$^{1}$\thanks{This work was supported by the Fonds québécois de la recherche sur la nature et les technologies (FRQNT) and the Natural Sciences and Engineering Research Council of Canada (NSERC).}\thanks{$^{1}$All authors are with the Department of Mechanical Engineering, Université de Sherbrooke, Qc, Canada.}\thanks{$^{2}$© 2022 IEEE. Personal use of this material is permitted. Permission from IEEE must be obtained for all other uses, in any current or future media, including reprinting/republishing this material for advertising or promotional purposes, creating new collective works, for resale or redistribution to servers or lists, or reuse of any copyrighted component of this work in other works.}}

\begin{document}

\maketitle

\begin{abstract}

Wearable robots are limited by their actuators performances because they must bear the weight of their own power system and energy source. This paper explores the idea of leveraging hybrid modes to meet multiple operating points with a lightweight and efficient system by using hydraulic valves to dynamically reconfigure the connections of a hydrostatic actuator. The analyzed opportunities consist in 1) switching between a highly geared power source or a fast power source, 2) dynamically connecting an energy accumulator and 3) using a locking mechanism for holding. Based on a knee exoskeleton case study analysis, results show that switching between gearing ratio can lead to a lighter and more efficient actuator. Also, results show that using an accumulator to provide a preload continuous force has great mass-saving potential, but does not reduce mass significantly if used as a power booster for short transients. Finally, using a locking valve can slightly reduce battery mass if the work cycle includes frequent stops. The operating principles of the proposed multimodal schemes are demonstrated with a one-DOF prototype.
\end{abstract}


\section{Introduction}

Wearable robots face issues similar to vehicle powertrains: the mass of the system and its energy efficiency are very important since the power source and energy source are on-board. However, as opposed to vehicle powertrains, the large majority of robotics systems rely on a single mechanical configuration to fit all operating modes. This often leads to oversized motors and poor efficiency. Automobiles for instance take advantage of multiple transmission ratios, independent passive brakes and a neutral point in the transmission in order to minimize the engine size and the energy consumption. 

This paper explores the potential gains of mass and efficiency when leveraging hybrid mechanical modes in a hydrostatic actuation system for wearable robots. The analyzed solutions are based on simple valves reconfiguring a hydrostatic transmission dynamically. The following topics discuss the litterature background relating to multimodal actuation systems and hydrostatic transmissions.

\subsection{Multimodal Actuation and Hydrostatic Transmissions}

\begin{figure}[t]
\centering
\includegraphics[width=0.48\textwidth]{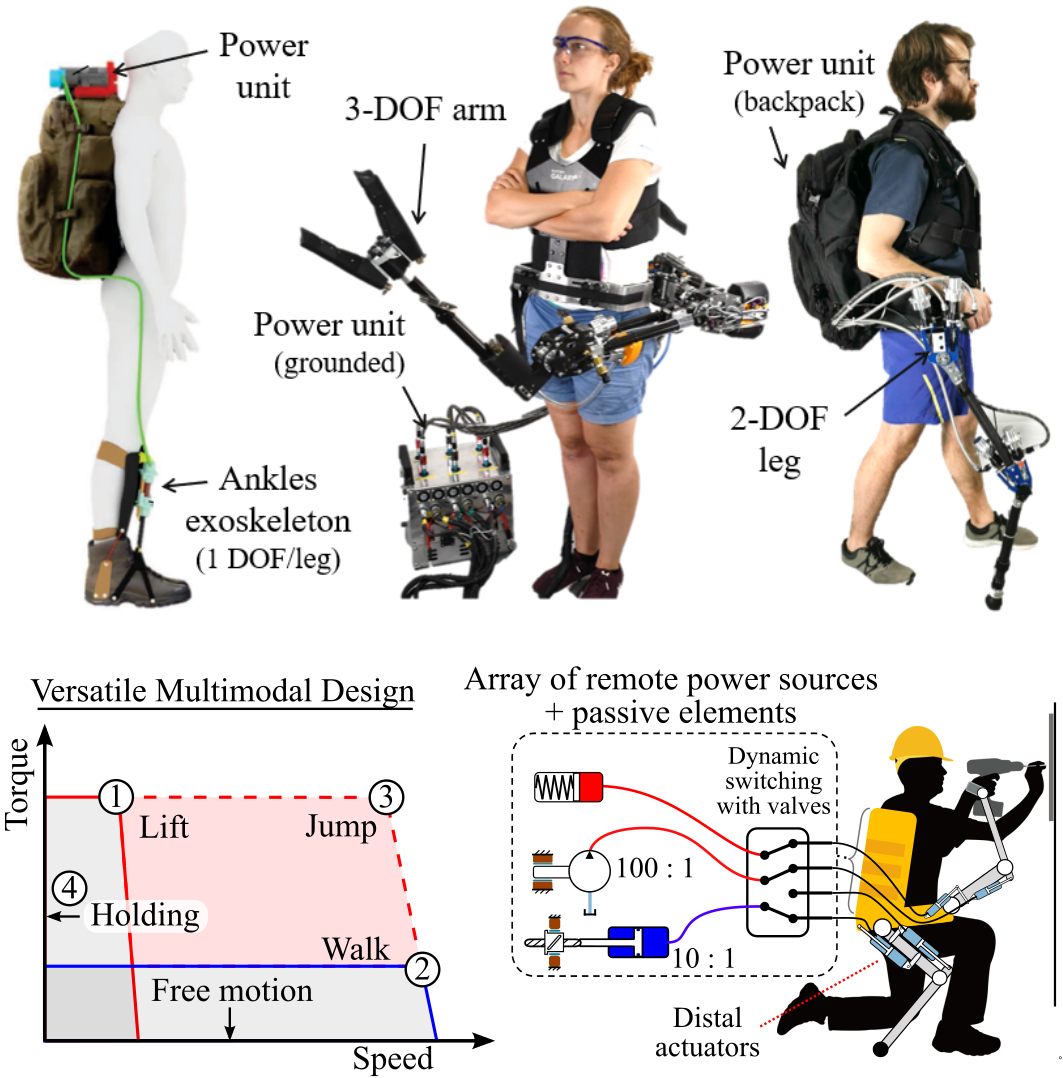}
\vspace{-10pt}
\caption{Examples of previous hydrostatic actuation units developed by the authors' laboratory (top) \cite{khazoom_design_2019} \cite{veronneau_multifunctional_2020} \cite{khazoom_supernumerary_2020} and a proposed multimodal power unit for multi-DOF wearable robots (bottom).}
\vspace{-10pt}
\label{fig_intro}
\end{figure}

\textbf{Variable transmissions} have been studied extensively for automobile powertrains but much less in the context of robot actuators. A few instances of research in that direction were made for systems facing bimodal operations where the robot has a phase at high speed and low force and then a phase at low speed and high force:
two-speed designs with one motor~\cite{phlernjai_jam-free_2017} \cite{lee_flexible_2019} and dual-motor designs for seamless gear shifting and efficiency~\cite{lee_new_2012,girard_two-speed_2015,verstraten_modeling_2018,jeong_design_2017}. Specifically, Lee \textit{et al.} made a lower limb exoskeleton with a dual reduction gearbox for switching between walking and sit-to-stand modes by using a bistable mechanism only at rest \cite{lee_flexible_2019}.

\textbf{Energy accumulators that can be disconnected} have been proposed in the context of robot actuators for storing and releasing energy efficiency, for instance in legged locomotion leveraging the cyclic motion \cite{rezazadeh_robot_2018}. Another possible use of energy accumulators is to provide a power boost as they are usually much more power-dense than electric motors \cite{midgley_comparison_2012}. Tsagarakis~\textit{et al.} proposed a parallel elastic antagonistic configuration for legged robots using a motorized parallel spring and a latching mechanism \cite{tsagarakis_asymmetric_2013}. Leach~\textit{et al.} presented a multimodal actuator with eight discrete modes such as energy storing and releasing, compliant or stiff transmission and rigid or free to move output \cite{leach_linear_2014}.

\textbf{Locking mechanisms and brakes} have been proposed as an energy-saving approach for robots that must bear forces while holding a position. Industrial robots are equipped with normally closed brakes that can reduce significantly power consumption during production-free time \cite{meike_energy_2011}. For transparent wearable robots, they require even more power to sustain a gravity load since their transmission is backdrivable.

\textbf{Hydrostatic transmissions} have been proposed for wearable robots for placing power sources remotely in a backpack, as opposed to directly on distal joints where the weight is highly burdensome for a user \cite{skinner_ankle_1990}. Fig. \ref{fig_intro} shows a few examples of explored applications by authors' laboratory: an exoskeleton, a wearable manipulator and a robotic cane. The advantage of hydrostatic transmissions are mainly: 1) easy routing through complex kinematics \cite{khazoom_design_2019}, 2) good force density and 3) allowing good backdrivability \cite{whitney_hybrid_2016}. Specifically, Sugihara~\textit{et~al.} proposed a reconfigurable hydrostatic circuit to allow a variable ratio using fast hydraulic switching with extra valves and cylinders. Much remains to be explored regarding this kind of multimodal hydrostatic approach.

In the previous literature, many hybrid actuation solutions have been proposed for many contexts. It is however hard for designers to extract guidelines regarding when a multimodal solution is worth the additional complexity for a gain in performance. The main contribution of this paper is a mapping of the trade-offs for mass and efficiency through the analysis of some multimodal hydrostatic solutions.

The paper is organized as follows:Section~\ref{sec:case} and~\ref{section_analyzes} presents the mass and energy gain potential of multimodal hydrostatic configurations to achieve extreme operating points for the case study of an knee exoskeleton. Section~\ref{section_Discussion} discusses the limitations of the analysis and results. Experimental proof-of-concept demonstrations of the proposed hybrid modes are presented in the video attachment and an appendix includes the modeling approach for estimating the mass of the components involved.\\





\section{A Multifunctional Exoskeleton Case Study}
\label{sec:case} 


This section presents sets of requirements and a baseline actuator that are used in the following section for comparing multimodal actuation solutions. Three tasks and associated requirements inspired by the challenges of actuating a knee exoskeleton are used:
\begin{itemize}
    \item Task 1: Lifting and walking with heavy payloads requiring high torque at low speeds;
    \item Task 2: Natural assisted walking requiring high speeds;
    \item Task 3: High power jumping for short transients.
\end{itemize}

Furthermore, in many situations, wearable robots must be transparent: they must not restrain the natural motion of the human wearer. This constraint on backdrivability, important for tasks 2 and 3, is simplified here into a constraint on reflected inertia from motorization. A maximum reflected inertia of 0.035~\si{\kg\meter\squared} is used based on a one-order-of-magnitude (10\%) smaller inertia than a 74~\si{kg} average man lower leg (calf and feet) inertia about knee axis \cite{ramachandran_estimation_2016}.

As illustrated by Fig.~\ref{fig_intro}, task requirements can be very different for torque and speed. Designing a robot that could be used for all these operating points is not usual. The Sarcos Guardian XO full-body exoskeleton can amplify human loading capacity but could not be used for natural walking and jumping \cite{sarcos_robotics_expanding_2020}. In contrast, Harvard's cable-driven soft exosuit can assist for loaded walking \cite{panizzolo_biologically-inspired_2016} but is not designed to increase human lifting or jumping capabilities significantly.

The parameters used for designing a knee exoskeleton that can do all tasks are given in Table~\ref{table_KneeParameters}. The 100~\si{\newton\meter} maximum torque is based on a human knee for a sit-to-stand motion from a low chair \cite{kaminaga_development_2010} so that lifting capabilities are increased significantly when using the exoskeleton. Maximum peak velocity is based on human walking speed increased by a factor of 1.5 to ensure natural motion at any time. The joint stroke is 160\si{\degree} based on maximal knee flexion \cite{lee_flexible_2019}.



\begin{table}[h!]
\renewcommand{\arraystretch}{1.1}
\caption{Knee exoskeleton parameters for comparison studies}
\label{table_KneeParameters}
\centering
\begin{tabular}{ l c c c }
\hline 
 & Torque $\tau_{\text{i}}$& Speed $\omega_\text{i}$ & Inertia $J_\text{i}$ \\
Task $i$ & (\si{\newton\meter}) & (\si{\per\second}) & (\si{\kg\per\meter\squared}) \\
\hline 
\hline
1) Lifting & 100$^\star$ & $9.4/\lambda$ & - \\
2) Walking & $100/\lambda$ & 9.4$^\star$ & 0.035$^\star$ \\
3) Jumping / high power & 100 & 9.4 & 0.035 \\
4) Holding & 100 & 0 & - \\
\hline
$^\star$Reference & \cite{kaminaga_development_2010} & \cite{lee_flexible_2019} & \cite{ramachandran_estimation_2016} \\
\hline
\end{tabular}
\end{table}


In order to evaluate the proposed multimodal solutions as a function of the exoskeleton target tasks and goals, a $\lambda$ variable is introduced in the requirements. The situation $\lambda=1$ describes a high-power exoskeleton that can provide high-lifting torque during fast motions continuously. Larger $\lambda$ represents a less powerful exoskeleton for which the lifting task is $\lambda$ times slower than the maximum walking speed. For instance, for the dual-speed exoskeleton for sitting-to-standing and walking in \cite{lee_flexible_2019}, the relative ratio is 2.8. 


\begin{figure}[h]
\centering
\includegraphics[width=0.48\textwidth]{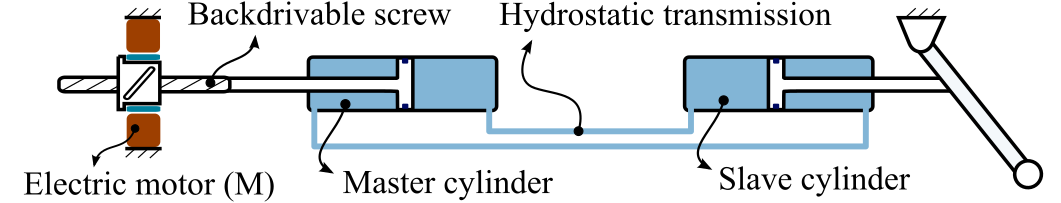}
\vspace{-10pt}
\caption{The baseline configuration used for analytical comparison.}
\vspace{-5pt}
\label{fig_BASELINE}
\end{figure}

Fig.~\ref{fig_BASELINE} shows the baseline solution consisting in a hydrostatic transmission, a high lead ball screw and an electric motor, which inspiration is drawn from~\cite{veronneau_multifunctional_2020,carney_design_2021}. In the next sections, new topologies are compared with the baseline one.

\section{Topologies Analysis}
\label{section_analyzes}
\subsection{Two Speed Switching}
\label{section_DSDM}

\begin{figure}[h]
\centering
\vspace{-10pt}
\subfloat[\centering]{{\includegraphics[width=0.19\textwidth]{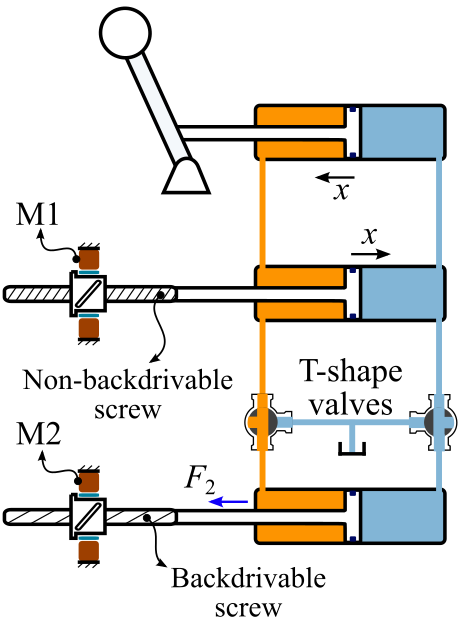}}}%
\qquad
\subfloat[\centering]{{\includegraphics[width=0.24\textwidth]{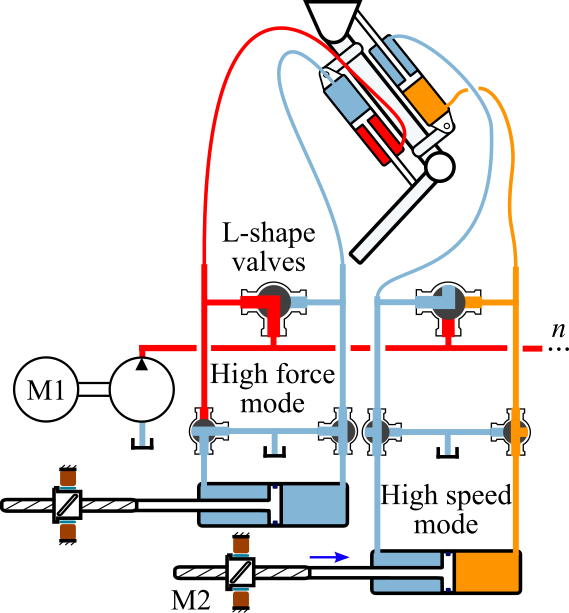}}}%
\caption{Two-speed configurations for a) 1 DOF using two master cylinders; b) $n$ DOF with a single high-pressure pump for all DOFs (here, $n=2$).}
\label{fig_DSDM}
\end{figure}

This section evaluates the idea of leveraging two different reduction ratios, i.e., a large reduction for slow lifting (task 1) and a small reduction for speed and backdrivability (task 2). A multimodal configuration (Fig.~\ref{fig_DSDM}(a)) and the baseline one (Fig.~\ref{fig_BASELINE}) are compared in terms of mass and efficiency when they are designed for meeting the requirements of both tasks. Furthermore, results are presented as a function of $\lambda$, a variable describing how far away in terms of torque and speed both tasks differ (Table~\ref{table_KneeParameters}). 

\begin{figure}[h]
\centering
\includegraphics[width=0.49\textwidth]{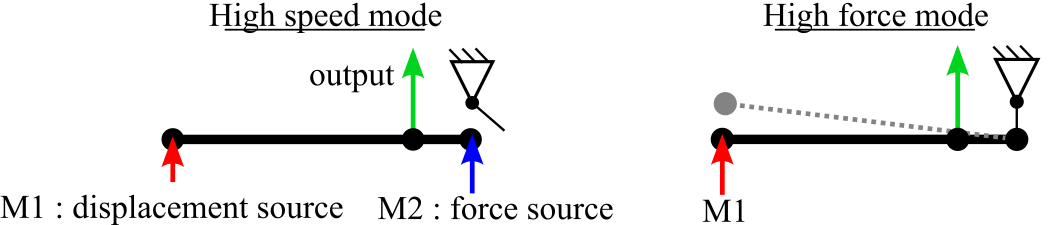}%
\vspace{-15pt}
\caption{Two modes of operation for two speed switching.}
\label{fig_DSDM_twoModes}
\vspace{-10pt}
\end{figure}

The multimodal solution in Fig.~\ref{fig_DSDM}(a) is a dual-motor configuration analogous to a previously proposed solution for addressing seamless dynamic gear shifts \cite{girard_two-speed_2015}, but here proposed in the hydraulic domain. In high-speed mode, the flow of both master pistons adds up and the pressure is shared when the valves are open, as shown in Fig.~\ref{fig_DSDM_twoModes}. This allows for a high-speed backdrivable mode with force capabilities limited by the weakest motor. For high force operation, the lightly geared motor (M2) is disconnected by closing its hydraulic connection, and the highly geared motor (M1) can drive the load alone at high force but limited speed.



\begin{figure}%
    \centering
    \subfloat[\centering]{{\includegraphics[width=0.22\textwidth]{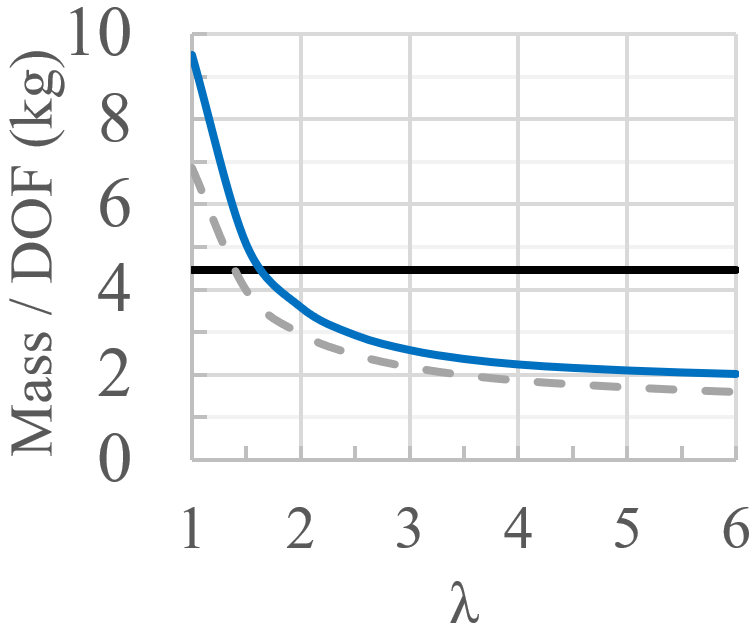}}}%
    \qquad
    \vspace{-10pt}
    \subfloat[\centering]{{\includegraphics[width=0.22\textwidth]{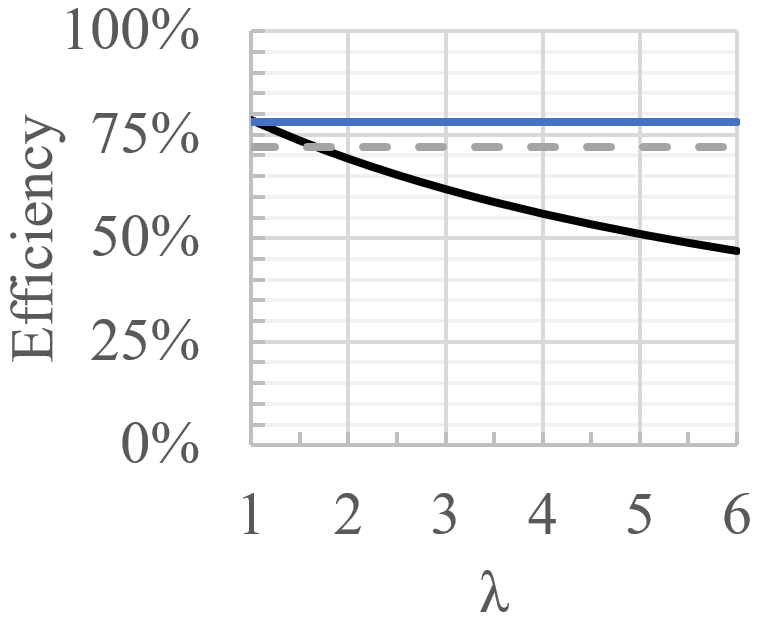}}}%
    \qquad
    \caption{Mass (kg/DOF) and power losses comparison of the baseline configuration (solid black) with the 1-DOF two-speed solution (solid blue) and the $2$-DOF solution with a shared pump (dashed grey).}%
    \label{fig_Results1_DSDM}%
    \vspace{-15pt}
\end{figure}

Fig.~\ref{fig_Results1_DSDM}(a) shows evaluated mass of both the multimodal and the baseline solution as a function of a ratio between operation points 1 and 2, defined as $\lambda={{\tau }_{1}}/{{\tau }_{2}}={{\omega }_{2}}/{{\omega }_{1}}$. The mass is found by first 1) computing the most advantageous reduction ratio for each configuration given the requirements. Then 2) computing the individual requirements (force, speed, power) of the components (piston, valves, ball screw and motors) present for each configuration. Finally, 3) using scaling laws for estimating the mass of all components. The approach, hypotheses and inputs are detailed in the Appendix. Results show that the two-speed solution is lighter for $\lambda>2$ ratios, i.e. when task 1 velocity is half the walking velocity. For $\lambda=3$, the multimodal solution would be 2.3~kg/DOF lighter, i.e., a 45\% mass reduction.

Regarding the energy efficiency, Fig.~\ref{fig_Results1_DSDM}(b) shows total efficiency of both solutions at operating point 1. The analysis considers Joule's losses only for motors, and a fixed ball screw efficiency. The baseline solution is at a disadvantage since the motor will be operating at a low inefficient velocity. For example, with $\lambda=3$ the energy for compensating the extra power loss of 2h operating at point 1 corresponds to 1.5~\si{\kg/DOF} of LiPos batteries (150~\si{\watt\hour} specific density). For the operating point 2, the energy efficiency of each solution is similar as motors operate close to their nominal velocity in each case ( $\approx 85\%$ for brushless DC motor).




Shown at Fig.~\ref{fig_DSDM}(b), another multimodal multiple-speed solution is evaluated. The idea is that the strong M1 and its master piston could be replaced by a hydraulic pump and, furthermore, a single pump could be used to power the lifting mode of multiple joints as it is the case in classical hydraulic designs. 
The 2-DOF multimodal design mass and efficiency are computed with the same approach. The pump's power rating is also set to provide the continuous power of task 1 for both DOF simultaneously and a fixed axial pump efficiency is used. The analysis assumes that on/off selection valves (as opposed to directional or servo valves in classical hydraulics) are used for this design, leading to a limitation that there would be coupling between DOFs when using the high force mode simultaneously. Results given in Fig.~\ref{fig_Results1_DSDM} shows that the 2-DOF design is slightly lighter than the first design even though more valves are needed. The energy-efficiency gain is not as high as for the 1-DOF solution because the pump is not as efficient as the ball screw.

\subsection{Accumulator}
\label{section_accumulator}
\begin{figure}%
 \centering
    \subfloat[\centering]{{\includegraphics[width=0.49\textwidth]{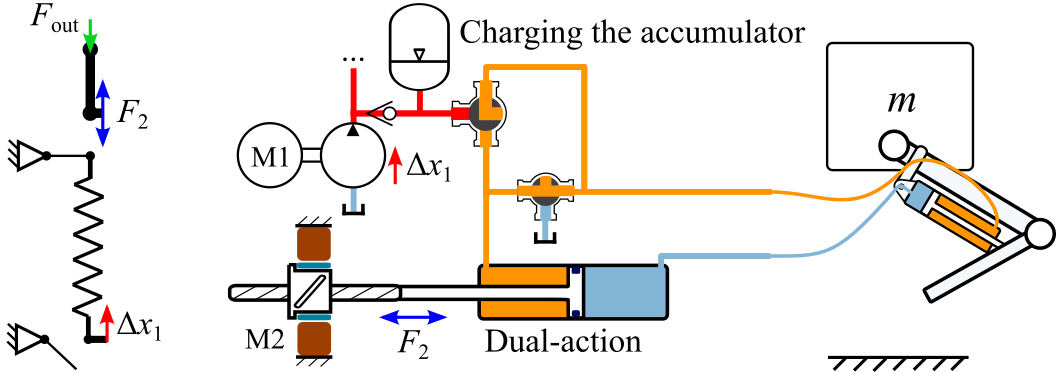}}}%
        \vspace{-12pt}
    \qquad
    \vspace{-3pt}
    \subfloat[\centering]{{\includegraphics[width=0.49\textwidth]{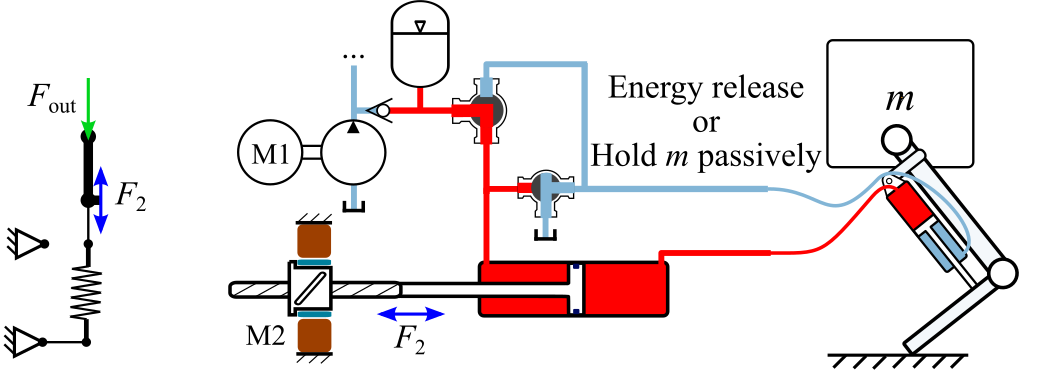}}}%
    
    \caption{A hydrostatic multimodal actuator with an accumulator driven by an auxiliary low power pump and acting in parallel on the slave actuator. a) Valves configuration for bidirectional actuation when no-load is applied (e.g., leg in swing phase) and for charging the accumulator. b) Valves configuration for power boosting or for providing a static force offset.}%
\label{fig_ACCUMULATOR}
\vspace{-5pt}
\end{figure}

This section analyzes the idea of adding an accumulator to a hydrostatic actuator for two goals. First, the idea of enabling intermittent \textbf{power boosting} capabilities (task 3) for a walking exoskeleton (task 2) is evaluated. Second, the idea of using the accumulator in parallel to passively \textbf{output a DC static force} (preload) that increases the continuous torque capability is explored too. A multimodal configuration (Fig.~\ref{fig_ACCUMULATOR}) is compared to the baseline one (Fig.~\ref{fig_BASELINE}) as a function of a $\lambda$ that reflects the ratio of torque assistance between task 2 (regular operation) and 3 (Table~\ref{table_KneeParameters}).

The proposed multimodal solution is composed of a high-speed backdrivable actuator M2, a large-volume hydraulic accumulator, a high-pressure miniature axial piston pump and two motorized custom ball valves for shifting between modes. The accumulator is connected in series with the pump so their flow adds up and pressure is shared. When valves are switched, the accumulator acts as a spring in parallel with the high-speed actuator so their forces add up. Left schematics in Fig.~\ref{fig_ACCUMULATOR} give an insight on how the spring unit can be connected and disconnected from the cylinder on purpose.




\subsubsection{Power Boosting}
For power boosting, the pump slowly fills the accumulator while the knee exoskeleton is controlled by the high-speed motor (Fig.~\ref{fig_ACCUMULATOR}(a)). The accumulated energy is released on request for task 3 and M2 can act in parallel to control the motion (Fig.~\ref{fig_ACCUMULATOR}(b)). If the power boosting is used for jumping, the accumulator could also be used to recover some potential energy during the landing.

\begin{figure}%
 \centering
    \vspace{-5pt}
    \subfloat[\centering]{{\includegraphics[width=0.22\textwidth]{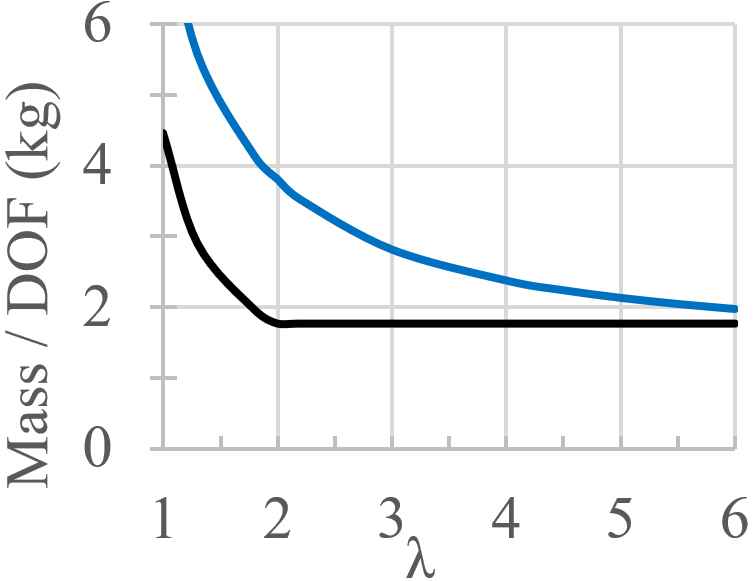}}}%
    \qquad
    \subfloat[\centering]{{\includegraphics[width=0.22\textwidth]{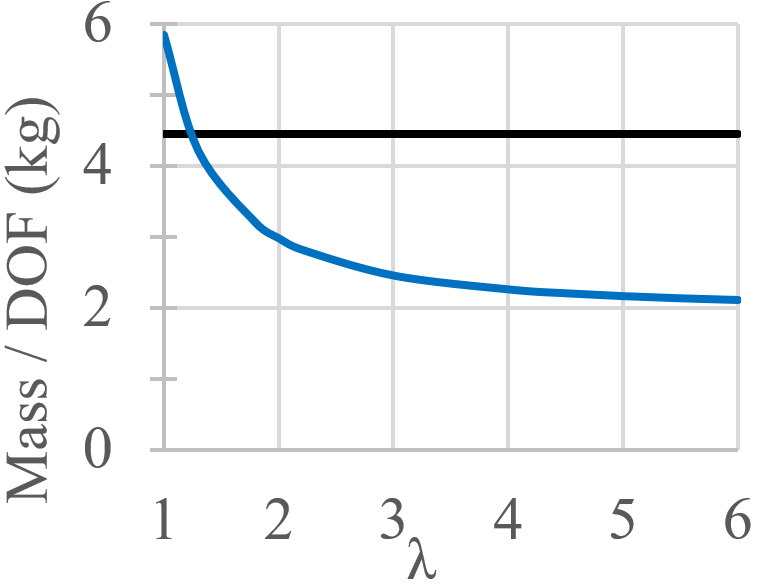}}}%
    \caption{Mass comparison of the multimodal configuration (solid blue) and the baseline one (solid black) for a) power boosting design and b) for providing a passive DC force design.}%
\label{fig_ScalingAccumulatorMass}
\vspace{-15pt}
\end{figure}

Fig.~\ref{fig_ScalingAccumulatorMass}(a) shows the mass for the multimodal solution and the baseline one, as a function of the ratio $\lambda$ representing how much more powerful the transient point 3 is compared to the continuous point 2 operation. The transient point 3 power is fixed ($\approx$ 942 w), and large $\lambda$ represents exoskeleton with less walking assistance. The mass of each configuration is computed based on a very similar methodology as in section~\ref{section_DSDM}: 1) find optimal reduction ratio, 2) compute individual components requirements and 3) compute components' mass using scaling laws. Motors peak torque is assumed to be twice the maximum continuous torque. For the multimodal solution, the pump is rated for the same continuous power as the high-speed line so that the peak power duty cycle is equal to $1/\lambda$. The accumulator volume is 100~\si{\milli\liter} which is five times the volume displaced by the slave piston so that the pressure drop is small during the power-boosting motion. Details regarding mass modeling are given in Appendix.

Results show that the baseline configuration is always lighter here due to the extra components of the multimodal solution. Actually, for the baseline design and $\lambda<2$, the peak power can be reached without increasing motor weight. For the multimodal solution, a mass saving is possible for large $\lambda$ and different requirements but it would be slim compared to the extra complexity and is thus probably not worth it in the presence of intermittent high power output.



\subsubsection{Parallel Static Force Offset}

The same hybrid architecture with the accumulator can also be used to provide a constant DC force passively, for instance to cancel out the mean gravity load of a system. Using a large-volume accumulator compared to the displaced volume in the cylinder (five times for the analysis, so 100~\si{\milli\liter}), the provided accumulator force would not vary much as a function of the cylinder motion. With this parallel configuration, the torque supplied by the accumulator unit adds up to the torque supplied by the high speed backdrivable motor M2. This configuration thus allows reaching the high force operating point 3 continuously, but with a bounded dynamic range for the torque level around the operating point. If the provided offset is larger than the maximum torque of M2, then the accumulator must be disconnected (as in Fig.~\ref{fig_ACCUMULATOR}a) to operate at low or negative torques (operating point 2). For instance, when a leg is in a swing phase, there is no gravity load to cancel out. 


Fig.~\ref{fig_ScalingAccumulatorMass}(b) gives the evolution of the computed mass for the multimodal and baseline solutions as the ratio $\lambda$ between operating points 2 and 3 increases. For this analysis, large $\lambda$ values represent an exoskeleton which the required dynamic torque range provided by M2 is $\lambda$ times smaller than the static torque offset. A maximum design pressure of 21~\si{\mega\pascal} is used to base the mass estimation on available commercial components. Pump unit power rating is arbitrary and would depend on how fast the force offset is required to change. For practical purposes, here it was fixed based on a requirement of filling the accumulator within 5~\si{\second}, giving a 230~\si{\watt} power rating for the pump unit. Mass is computed for all components based on individual requirements and scaling laws are detailed in the Appendix.



Results show a large mass gap between designs for $\lambda>1.5$. For $\lambda=3$, the mass saving is 2.1~\si{kg/DOF}. This multimodal solution is worth exploring for wearable robotic actuators often working against a mean gravity load. 
\subsection{Locking Valve}
\label{section_Locking_valve}

\begin{figure}[ht]
    \centering
    \includegraphics[width=0.49\textwidth]{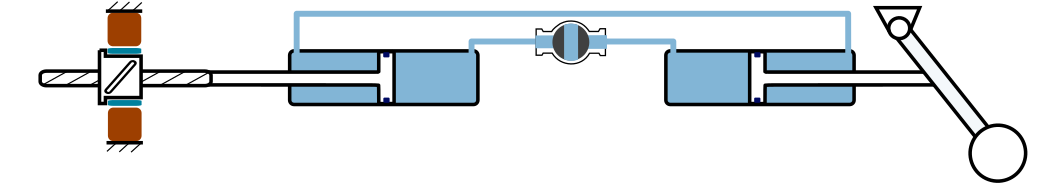}%
    
    \caption{A motorized locking 2/2 valve for a hydrostatic transmission.}%
     \vspace{-5pt}
    \label{fig_LOCKING}%
 
\end{figure}

This section leverages the idea of using a motorized ball valve for holding a large load passively at zero velocity by simply blocking the flow passing through a hydrostatic transmission (see Fig.~\ref{fig_LOCKING}). The energy efficiency of this solution is assessed in the context of a duty cycle between operating at point 3 and regular breaks for holding the load at zero velocity (operating point 4).

\begin{figure}%
    \centering
 \vspace{-10pt}
    \subfloat[\centering]{{\includegraphics[width=0.21\textwidth]{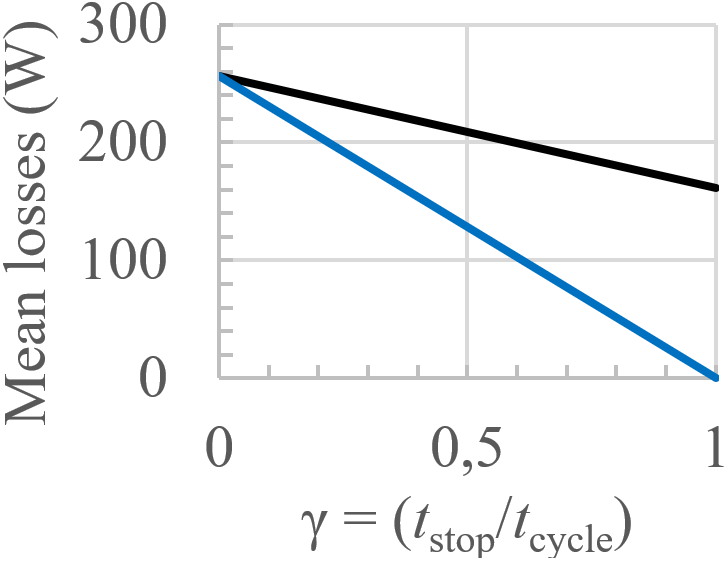}}}%
    \qquad
    \subfloat[\centering]{{\includegraphics[width=0.21\textwidth]{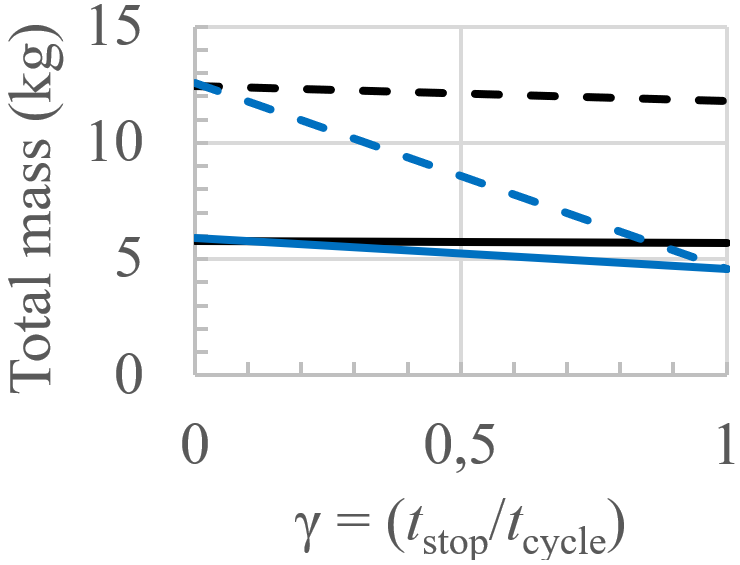}}}%
    \caption{Comparison of mean power losses (a) and total mass including batteries and the valve (b) for a high-power work cycle (task 3) with holding breaks (task 4). Extra battery mass is for a 10~\si{\min} (solid) and a 1~\si{\hour} (dashed) autonomy. Baseline configuration is in black and the multimodal one with a locking valve is in blue.}%
    \label{fig_Results_Dissipative}%
    \vspace{-10pt}
\end{figure}

Power losses and equivalent battery mass are computed and presented in Fig.~\ref{fig_Results_Dissipative}. As opposed to previous analyses, the design here is fixed for both the baseline and the multimodal configuration, i.e., with same requirements for torque, speed and reflected inertia (see table~\ref{table_KneeParameters}). The multimodal configuration prevents operating at zero efficiency for holding during a portion of the cycle ${t}_{\text{stop}}=\gamma{t}_{\text{cycle}}$. At $\gamma=0$, the exoskeleton is always moving. At $\gamma=1$, the exoskeleton is always holding the load. Calculations include motor and ball screw losses. An equivalent mean efficiency through the work cycle is computed. Battery mass comparison is made for a 10~\si{\min} and a 1~\si{\hour} total cycle. Mass results show that the multimodal solution is advantageous for almost any $\gamma$ because the locking valve adds very little mass compared to the mass of batteries saved through minimized power losses. The mass-saving potential is, however, only significant for a long work cycle.


\section{Discussion}
\label{section_Discussion}
The feasibility of the multimodal solutions was explored with an experimental setup based on industrial components. A 1-DOF leg is actuated either by a high-speed actuator, a high-force actuator or a hydraulic accumulator. Demonstrations in the attached video show seamless shifting between high force, high speed and power modes. Still, this section discusses the limitations regarding the above analyses. 


\textbf{Mass analyses} do not include framing and electronics. A multimodal approach is more complex and probably requires a heavier and bulkier frame as well as more electronics. Also, some multimodal configurations would need extra hydraulic components for proper use, e.g., charge pump, reservoir, check valves, pressure relief valves, filter and manifold. Only pumps and ball valves mass were considered. Custom multi-port valves could reduce the number of valves. A preliminary complete design is recommended to give a better insight of the mass-saving potential of a given multimodal topology.


\textbf{High-pressure ratings} were used in this paper because optimized low-pressure pumps and accumulators are not commercially available. Working at lower pressures could degrade the validity of the multimodal solutions. Indeed, for a same work-per-stroke specification, more flow is necessary, which means: 1) more fluid volume in the transmission, accumulator and reservoir, 2) bigger hoses for same viscous losses and 3) bigger cylinders. The overhead of extra components would thus disadvantage multimodal solutions at lower operating pressures.

\textbf{Cylinder seal friction} could deteriorate the backdrivability of the high-speed mode since it is designed for high-pressure mode too. Indeed, the slave actuator is common to both modes but must withstand high pressure. Friction-less rolling diaphragm cylinders could be used to minimize friction but would affect negatively the overall system mass. 


\textbf{Sensitivity to input parameters} The exact threshold points, at which multimodal solutions become advantageous over the baseline one, were found to be sensitive to electric motor torque density and torque-to-inertia ratio as well as to the velocity of the operating points. For example, the mass advantage trend of the two speed switching solution would drop if electric motors were four times more torque-dense.
\section{Conclusion}
This paper explored the potential gain of adding hybrid modes to hydrostatic designs for wearable robotics. In the context of a knee exoskeleton, mass and energy analyses led to the following observed trends:

\begin{itemize}
    \item \textbf{Two-speed:} This approach is promising for mass and efficiency when high power is not a necessity. It is worth considering this design option when developing a multi-function wearable robot.
    \item \textbf{Accumulator for power boost:} This solution is probably not worth the additional complexity given slim potential mass savings.
    \item \textbf{Accumulator for static force offset:} There is a mass saving opportunity for high force wearable robots and mobile robots that must constantly fight a gravity load.
    \item \textbf{Locking valves:} A locking mechanism may be valuable if the work cycle includes frequent stops under load, and long autonomy is targeted.
\end{itemize}

The experimental setup highlighted some implementation challenges regarding friction, stroke limitations and complex bleeding and air leak prevention. Future work includes improving the hydraulic design and the development of control algorithms for coordinating seamless hydraulic gear shifting between the hybrid modes. The trends presented in this paper will also be used to develop actuation system for a specific application to validate if the potential mass saving can be realized experimentally.

\section*{Appendix}

\label{section_modeling}
This section presents the mass modeling of the main components in the designs. It is based on commercial components. Their properties are scaled using scaling laws which give a general trend of the main properties $y$ of system components based on geometric and materials similarity. The form is given by $y=kx^a$, where the reference parameter is $x$ and the scaling parameters are $k$ and $a$. Scaling laws were used for instance for designing complex mechatronic systems \cite{budinger_estimation_2012}, for motor geometry design \cite{seok_actuator_2012} and for designing gearmotors for robots~\cite{saerens_scaling_2019} \cite{saerens_scaling_2020}. Scaling parameters used here are summarized in Table~\ref{table_scalingParameters} with details in next subsections.


\begin{table}[h!]
\renewcommand{\arraystretch}{1.1}
\caption{Scaling law parameters used for modeling, $y=kx^a$}
\label{table_scalingParameters}
\centering
\begin{tabular}{ r c c c c c }
\hline
 \multicolumn{1}{l}{Component and} & Scaled & &  & & \\
 \multicolumn{1}{l}{property $y$} & from $x$ &  $k$ & $a$ & Units & Ref.\\
\hline 
\hline
\multicolumn{1}{l}{Electric motor (M)} & $\tau_\text{M}$ & & & \si{\newton\meter} & \cite{technotion_frameless_2019} \\ 
Mass & & 0.30 & 0.71 & \si{\kg} &  \\
Nominal speed & & 309 & -0.64 & \si{\per\s} &  \\
Rotor inertia & & 2.1E-05 & 1.42 & \si{\kg\m\squared} &  \\
\hline
\multicolumn{1}{l}{Ball screw (BS)} & $F_\text{BS}$ & & & \si{\newton} & \cite{nsk_nsk_2013} \\ 
Force density & & 15000 & 0 & \si{\newton\per\kg} &  \\
\hline
\multicolumn{1}{l}{Accumulator} & $V_\text{displac.}$ & & & \si{\liter} & \multicolumn{1}{l}{\cite{mallick_enabling_2017}} \\ 
Mass & & 0.95 & 0.56 & \si{\kg} & \multicolumn{1}{l}{\cite{tam_design_2000}} \\
\hline
\multicolumn{1}{l}{Axial piston pump} & $P_\text{nom}$ & & & \si{\watt} &  \cite{takako_small_nodate} \\ 
Power density & & 133 & 0.30 & \si{\watt\per\kg} &  \\
\hline
\vspace{-20pt}
\end{tabular}
\end{table}

\subsubsection{Reduction ratio selection}
The selection of the best reduction ratio $N$ for an application depends on the desired torque, speed and reflected inertia from motor. Inertia is amplified by $N^{2}$ through the transmission, $J_i=J_\text{M}N^2$. Then, the maximum ratio is either limited by motor inertia $J_\text{M}$ or by motor nominal speed $\omega_\text{M}$: 

\begin{equation}
    N=\min \left\{ \begin{matrix}
   \sqrt{{{{J}_{\text{i}}}}/{{{J}_{\text{M}}}}\;}  \\
   {{{\omega }_{\text{M}}}}/{{{\omega }_{\text{i}}}}\; 
\end{matrix} \right.
\label{eq_RatioInertiaSpeed}
\end{equation}

\begin{equation}
     {{\tau }_{\text{M}}}={{{\tau }_{\text{i}}}}/{(\eta_\text{BS}{N})} 
    \label{eq_RatioTorque}
\end{equation}

where $J_\text{M}$ and $\omega_\text{M}$ are scaling functions from Table~\ref{table_scalingParameters} depending on motor torque $\tau_\text{M}$ which also depends on $N$, on the transmission efficiency and on the joint torque. From equations~\ref{eq_RatioInertiaSpeed} and~\ref{eq_RatioTorque}, the unknowns $N$ and $\tau_\text{M}$ can be found from the output torque, speed and inertia requirements.


\subsubsection{Motors}
When coupled with lightly geared transmissions, brushless DC in-runner motors with large diameters are torque dense with reasonable reflected inertia \cite{seok_actuator_2012}. In this paper, the frameless torque motor series from Tecnotion is used as a reference (QTR-A-105-34Z) \cite{technotion_frameless_2019}. Maximum speed is based on a 48~V voltage supply. The specified ultimate torque from the datasheet is twice the rated continuous torque, which is used as the peak torque limit.





\subsubsection{Ball screws}
Ball screws are torque dense rotational-to-linear mechanisms with high forward and reverse efficiency ($\eta_\text{BS}>$~90\% for lead angles higher than 5°) \cite{thk_features_nodate} \cite{nsk_nsk_2013}. They can achieve a wide reduction range, have low backlash and can transmit both compression and tensile forces. 
%
%
For scaling small stroke assemblies, screw and nut mass must be considered, as well as extra screw length due to the nut. Scaling laws were fit on NSK ball screws (RNFTL model, \cite{nsk_nsk_2013}) for a 50~\si{\milli\meter} stroke. The total force density found is about constant with 15~\si{\kilo\newton\per\kg} for a 0.5--15~\si{\kilo\newton} range.




\subsubsection{Accumulators}
Hydraulic accumulators store energy in the form of compressed gas. Some composites accumulators were developed for mobile applications and are about five times more power dense than steel counterparts \cite{midgley_comparison_2012}.
%
%
%
%
SteelHead Composites diaphragm accumulators (MicroForce serie) are used here as a reference (smaller model is 24~\si{\mega\pascal} rated, 0.5 L nominal, 0.45 \si{\kg} \cite{mallick_enabling_2017}). The serie only includes two models so a research composite tank \cite{tam_design_2000} is added to data for fitting a scaling law based on maximum displaced volume. The accumulator is considered here as an ideal spring (100\% efficient) and maximum compression ratio is~6.


\subsubsection{Pumps}
Takako's miniature axial piston pumps were used in several mobile robotics projects \cite{khan_development_2015} and are used as reference here. Working pressure is up to 21~\si{\mega\pascal} and maximum power output is 0.2--6.5~\si{\kilo\watt} \cite{takako_small_nodate}. Mass is scaled as a function of power and excludes the addition of valves, fittings, filters and other hydraulic components necessary for proper use of the pump. Efficiency used for pump sizing is 80\%, a conservative value for axial pumps \cite{nath_optimization_2017}.


 


\subsubsection{Cylinders}
A 21~\si{\mega\pascal} actuator is chosen since available axial pumps and accumulators are designed for this pressure or over. For same work per stroke, high-pressure cylinders are lighter and less flow is required, so smaller hoses can be used. For the desired torque and stroke from Table~\ref{table_KneeParameters}, a 50~\si{\milli\meter} stroke cylinder KNR LSS05-0050 is selected. For a 160° angular stroke for the knee, a 18~mm effective radius of action is thus required, meaning that 100~Nm is achieved when the piston is generating 5500~N. This is the maximum retracting force of this cylinder. According to the manufacturer, the unit empty mass is 0.56~\si{kg}. All cylinders dimensions are fixed for simplicity.



\subsubsection{Motorized ball valves}
The mass of a custom high-speed motorized ball valve is estimated here. Ball valves are chosen for their mechanical simplicity and their low pressure drop. Valve mass comprises 1) the valve body, 2) the motor and 3) the gearbox.
The body is modeled as a 6.3~\si{\milli\meter} inner diameter hollow cylinder under pressure and made from grade 7075 aluminum. The volume and body mass are estimated by calculating the wall thickness based on the burst tensile stress and a safety factor of two. A correction factor based on commercial ball valves rated at 21~\si{\mega\pascal} is applied to account for the mass of other internal components, leading to an estimated valve body mass of 47~\si{\gram}.
%
%
%
The motor and gearbox estimated mass is based on the required opening speed as well as the breakaway torque. The latter is based on the estimated seat friction caused by the pressure acting on the ball. The following specifications were used: a torque of 1~\si{\newton\meter} and an opening time of 50~\si{\milli\second}, leading to a 30~\si{\watt} required power. A working combination of commercially available motor and gearbox (KDE2306XF-2050 motor and the Maxon GPX 22 planetary gearbox 3.9:1 ratio) leads to a total estimated mass of 138 g.


\subsubsection{Example}
A power unit mass calculation of the two-speed switching analysis (Fig.~\ref{fig_Results1_DSDM}a) is given here as an example. For any $\lambda$ of the baseline solution, the target speed and torque are 9.4~rad/s and 100~Nm. Based on Table~II parameters and equations~\ref{eq_RatioInertiaSpeed} and~\ref{eq_RatioTorque}, the maximum reduction ratio is 3.5 for the inertia and 3.8 for the speed, so $N=3.5$ and $\tau_\text{M}=(100~\text{Nm})/((0.9)(3.5))$=32~Nm. Also, a 5500~N ball screw force $F_\text{BS}$ must be applied on the piston to achieve to high torque point. The total power unit mass is then:

\begin{equation}
   {{m}_{\text{total}}}=\underbrace{{{k}_{\text{M}}}{{\tau }_{\text{M}}}^{{{a}_{\text{M}}}}}_{\text{motor}}+\underbrace{\frac{{{F}_{\text{BS}}}}{{{k}_{\text{BS}}}{{F}_{\text{BS}}}^{{{a}_{\text{BS}}}}}}_{\text{ball screw}}+\underbrace{{{m}_{\text{cyl}}}}_{\text{cylinder}} \\
 \end{equation}
 \begin{equation}
     {{m}_{\text{total}}}=\underbrace{0.30{{(32)}^{0.71}}}_{\text{motor}}+\underbrace{\frac{5500}{15000{{(5500)}^{0}}}}_{\text{ball screw}}+\underbrace{0.56}_{\text{cylinder}} \\ 
 \end{equation}

The total mass is then ${{m}_{\text{total}}}=3.5+0.37+0.56=4.43\text{ kg}$.




\bibliographystyle{IEEEtran}
\bibliography{Zotero}

\end{document}